\documentclass[letterpaper]{article} 
\usepackage{aaai2026}  
\usepackage{times}  
\usepackage{helvet}  
\usepackage{courier}  
\usepackage[hyphens]{url}  
\usepackage{graphicx} 
\usepackage{natbib}  
\usepackage{caption} 
\usepackage{algorithm}
\usepackage{algorithmic}

\usepackage{booktabs} 
\usepackage{enumitem}

\newcommand{\keypoint}[1]{\noindent \textbf{#1}\quad}

\usepackage{amsmath}
\usepackage{amssymb}
\usepackage{mathtools}
\usepackage{amsthm}
\usepackage{xcolor}

\newcommand{\ours}{Gen1S}
\DeclareMathOperator*{\argmax}{argmax} 

\usepackage{newfloat}
\usepackage{listings}
\DeclareCaptionStyle{ruled}{labelfont=normalfont,labelsep=colon,strut=off} 
\floatstyle{ruled}
\newfloat{listing}{tb}{lst}{}
\floatname{listing}{Listing}
\nocopyright 

\title{Feature-Space Generative Models for One-Shot Class-Incremental Learning}
\author {
    Jack Foster, 
    Kirill Paramonov, 
    Mete Ozay, 
    Umberto Michieli
}
\affiliations {
    Samsung R\&D Institute UK, United Kingdom
}

\begin{document}

\maketitle

\begin{abstract}
Few-shot class-incremental learning (FSCIL) is a paradigm where a model, initially trained on a dataset of base classes, must adapt to an expanding problem space by recognizing novel classes with limited data. We focus on the challenging FSCIL setup where a model receives only a single sample (1-shot) for each novel class and no further training or model alterations are allowed after the base training phase. This makes generalization to novel classes particularly difficult. We propose a novel approach predicated on the hypothesis that base and novel class embeddings have structural similarity. We map the original embedding space into a residual space by subtracting the class prototype (i.e., the average class embedding) of input samples. Then, we leverage generative modeling with VAE or diffusion models to learn the multi-modal distribution of residuals over the base classes, and we use this as a valuable structural prior to improve recognition of novel classes. 
Our approach, \ours, consistently improves novel class recognition over the state of the art across multiple benchmarks and backbone architectures.
\end{abstract}

\section{Introduction}
\label{sec:intro}


Deep neural networks have become increasingly ubiquitous; however, their reliance on training with independent and identically distributed (i.i.d.) data makes them vulnerable to catastrophic forgetting when exposed to non-i.i.d.\ data streams, which are common in real-world scenarios. For instance, a deployed robot must continuously adapt to recognize new objects as the environment evolves. Class-Incremental Learning (CIL) aims to address such settings by allowing models to sequentially learn novel classes while retaining their ability to classify previously learned ones. Unlike conventional offline supervised learning, where complete data coverage is assumed, real-world scenarios often involve limited data availability. For example, users may be unwilling or unable to provide multiple training samples. Few-Shot Class-Incremental Learning (FSCIL) extends the CIL paradigm to address this scarcity, requiring models to learn novel classes with only a handful of samples.

Nonetheless, existing FSCIL methods typically assume access to at least 5–10 samples per novel class, sufficient for meaningful updates to model parameters \cite{zhao2023few} or constructing reliable approximations of class distributions \cite{liu2023learnable,hayes2020lifelong}. 
These updates are often accompanied by computationally-demanding knowledge distillation training to maintain performance on previously learned classes \cite{li2017learning, cheraghian2021semantic, zhao2023few,michieli2019incremental}. 
In this work, we explore a more challenging scenario, where only a single exemplar is available for each novel class (one shot) and on-device training is infeasible. This reflects real-world deployment scenarios on resource-constrained devices, where models must adapt seamlessly using a single user-provided example. We refer to this setting as one-shot class-incremental learning (1SCIL).

\begin{figure}[tbp]
    \centerline{
        \includegraphics[width=1\linewidth]{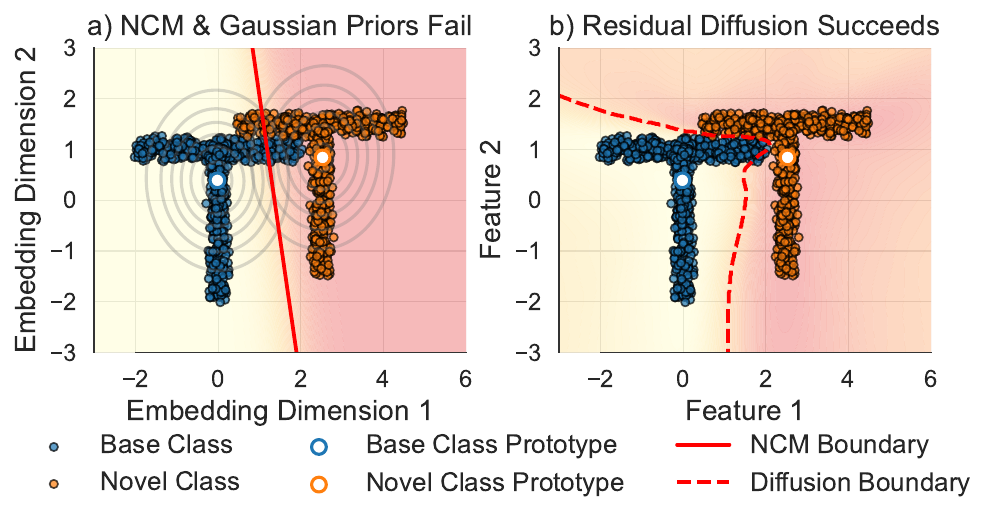}
    }

    \caption{One-shot generalisation via diffusion model. 
    Left: both nearest-class-mean and Gaussian classifiers collapse each T‑shaped class embeddings into a single mode, forcing a linear boundary (red solid line) that slices through the true clusters. 
    Right: a diffusion model trained only on the residuals of the base class captures the multimodal geometry and generalizes to the novel class, yielding a nonlinear decision boundary (red dashed line) that separates the two clusters despite observing just a single sample from the novel class.}
    \label{fig:setup}
\end{figure}

Under these stringent constraints, many existing methods struggle due to: a) their need for multiple labelled exemplars for each novel class to construct meaningful representations \cite{snell2017prototypical}; or b) their incorrect assumption that embedding distributions are unimodal or Gaussian \cite{hayes2020lifelong,liu2023learnable,snell2017prototypical, song2023learning}.
To overcome these limitations, we introduce a novel method for 1SCIL that we name \ours. Our method integrates a generative model that captures the shared embedding structure of base classes and extends it over novel classes to build a strong prior for classification of novel class samples.
Our contributions are motivated by two key observations from prior work:
(i) semantically related classes exhibit similar structures in the embedding space of a well-trained and a well-generalized backbone
\cite{zhang2018visual,kornblith2019better}, and
(ii) in FSCIL, the embeddings of novel classes may be approximated by mixing base class embeddings \cite{zhang2015zero}.
Additionally, our empirical findings reveal that embedding distributions are often multi-modal and complex, violating the Gaussian assumption common in prior FSCIL works.

%
To exploit this, we map the original embedding space onto a residual space by subtracting the respective class prototype. 
Then, we employ a generative model to learn the single shared distribution of base class residuals and predict it over the single labeled novel class sample.
By leveraging this shared structure as a prior for novel class structure, our method achieves robust one-shot learning performance without the need for multiple exemplars, addressing a critical gap in current methods (see Fig.~\ref{fig:setup}).
To the best of our knowledge, we are the first to apply generative models in the context of FSCIL. 
Unlike traditional approaches, our method can handle arbitrarily complex, multi-modal distributions learned from base class embeddings, enabling more effective generalization to novel classes.

In summary, our contributions are as follows:
\begin{itemize}[noitemsep,topsep=0pt]
    \item We observe that base and novel class embeddings in FSCIL setups exhibit similar intra-class structures, i.e., distributions of base and novel class residuals are similar.
    \item We introduce \ours, the first FSCIL method to leverage a generative model (e.g., VAE or diffusion models) in the embedding space. Trained on the base class residual distributions, our generative model serves as a prior to predict novel class distributions. 
    \item We achieve state-of-the-art results in the challenging 1SCIL setup, excelling in novel class recognition while maintaining the performance for base classes.
\end{itemize}

\section{Related Work}\label{related_work}
\keypoint{Few-Shot Class Incremental Learning}
FSCIL is a paradigm in which a model incrementally learns to recognize novel classes from a small set of samples while preserving previous knowledge \cite{tao2020few}. FSCIL approaches include replay- and distillation-based methods \cite{kukleva2021generalized, zhao2023few}, dynamic structure techniques \cite{zhang2021few}, meta-learning strategies \cite{hersche2022constrained, mazumder2021few}, and metric learning approaches \cite{mensink2013distance, snell2017prototypical}. 
Recently, \citealp{park2024pre,paramonov2024swiss} have also shown that pre-trained vision transformers are effective for FSCIL.

This work focuses on metric learning strategies, which exploit the feature space to enable rapid adaptation to new classes while mitigating forgetting of prior knowledge. 
Nearest-class-mean (NCM) classifiers applied in embedding spaces have shown effectiveness \cite{mensink2013distance}. Building on NCM, methods like Prototypical Networks \cite{snell2017prototypical} optimize feature convergence within classes and divergence across classes using contrastive loss. Extensions like FACT \cite{zhou2022forward} and SAVC \cite{song2023learning} generate virtual prototypes or classes to reserve embedding space for future novel classes, while LIMIT \cite{zhou2023few} synthesizes auxiliary tasks. NC-FSCIL \cite{yang2022neural} aligns prototypes to simplex equiangular tight frames based on neural collapse theory. OrCO \cite{ahmed2024orco} employs supervised and self-supervised contrastive learning to disentangle embeddings. RelationNet \cite{sung2018learning} uses a regression head on backbone embeddings to compute pairwise similarity scores.
On the other hand, SLDA \cite{hayes2020lifelong} improves NCM by modeling each class as a Gaussian distribution with a shared covariance across all classes that is initialized over base class samples and updated online on novel samples. Another version, SLDA-$\Sigma$-fixed, computes the shared covariance during pre-training and then keeps it fixed.
SimpleShot \cite{wang2019simpleshot} combines an NCM classifier with mean subtraction and L2-normalization. Similarly, \citealp{zhang2019variational} introduce a variational Bayesian framework to perform FSCIL. Using a Gaussian prior, \citealp{yang2021free} use statistics from classes with abundant data to inform the fitting of classes with fewer samples.


Although these methods effectively avoid extensive model updates when learning novel classes via nearest neighbor classification, they still require several samples to produce good estimates of novel class embeddings or covariance. 
Furthermore, they often rely on data- and computationally-expensive knowledge distillation to recover performance on base classes.
Finally,
they
rely on a Gaussian prior for novel classes \cite{snell2017prototypical,sung2018learning,hayes2020lifelong, yang2021free}.
In contrast, our approach enhances flexibility and informativeness by learning a prior directly from base classes and does not require training when learning new ones.


\keypoint{Generative Models}
Generative models are becoming increasingly popular and widely used in various applications.
Variational Autoencoders (VAEs) \cite{kingma2013auto}, which compress input data into a continuous probabilistic latent representation before reconstructing it with a decoder.
Generative Adversarial Networks (GANs) \cite{goodfellow2014generative} employ a generator to produce samples from a noise vector and a discriminator to differentiate between real and generated samples. 
Diffusion models \cite{ho2020denoising} map noise sampled from a Gaussian distribution to a target one through multiple iterative denoising timesteps. 
A relevant work \cite{bordes2022high} conditions a diffusion model on embeddings of image models to visualize their representations. 

Generative models have also found applications in classification \cite{ng2001discriminative, li2023your}, where a sample is generated under class conditioning, and the class that minimizes the error between generated and query test sample is selected as the output prediction. 
In traditional CIL, VAEs have been employed \cite{van2021class} where abundant data enables training a new VAE for each task.
Another common application is generative replay \cite{shin2017continual,maracani2021recall}, where the distribution of past classes is captured by a generative model to allow retraining on generated samples to mitigate forgetting. \citealp{xu2022generating} use a conditional VAE to generate samples for sparse classes in FSCIL, whereas \citealp{schwartz2018delta} leverage an autoencoder to extract transferable intra-class transformations between base and novel classes to synthesize data for few-shot object recognition. Within the FSCIL context, \citealp{zhang2018metagan} introduce an adversarial approach based on GANs, while \citealp{du2023protodiff} pairs meta-learning with a diffusion model to reduce the bias in few-shot prototypes that are later used with an NCM classifier.

Unlike these methods, we address the 1SCIL setup with no capability to train a model at deployment. Instead, we leverage generative models to learn informative priors over base classes that are used to improve novel class recognition.

\section{Problem Setup}
\label{sec:problem_setup}

\newenvironment{redtxt}{\par\color{red}}{\par}

We consider a set of classes $\mathcal{K}$ where each class ${k\in\mathcal{K}}$ is associated to a set of sample-label dataset pairs. We define:
\begin{itemize}[noitemsep,topsep=0pt]
    \item \textbf{Base classes:} The set of classes $\mathcal{K}_{b} \subset \mathcal{K}$ available at pre-training, before the model has been deployed, and for which there is abundant training data. 
    The dataset associated to $\mathcal{K}_{b}$ is split randomly into train and test sets, $\mathcal{D}^{(b)}_{train}=\{X^{(b)}_{train}, Y^{(b)}_{train}\} \text{ and } \mathcal{D}^{(b)}_{test}$, respectively.
    \item \textbf{Novel classes:} The set of classes $\mathcal{K}_{n} \subset \mathcal{K}$ available after pre-training on base classes, and for which data is scarce (with $\mathcal{K}_n \cap\mathcal{K}_b=\emptyset$).
    The dataset associated to $\mathcal{K}_{n}$ is split into support set $\mathcal{D}^{(n)}_{support}=\{X^{(n)}_{support}, Y^{(n)}_{support}\}$, used to set prototypes or to train the model, and a holdout query set $\mathcal{D}^{(n)}_{query}$ for evaluation.
\end{itemize}
Given a backbone model, $f(\cdot)$, optionally pre-trained on a broad dataset (e.g., ImageNet), an FSCIL pipeline typically begins with training $f(\cdot)$ to convergence on base class samples $\mathcal{D}_{train}^{(b)}$ at the server side. 
Then, $f(\cdot)$ is deployed on limited-resource devices (e.g., smartphones or robotic agents) and encounters novel class samples with scarce labeled data, requiring adaptation via an FSCIL method. 

In this paper, we focus on the challenging 1SCIL problem where each novel class has only a single annotated sample. Under such extreme data scarcity, making meaningful updates to the model becomes impractical.
To address this, we freeze the model parameters simplifying the learning process and eliminating the need for on-device training. 
Since the model is not being trained and the classification of novel classes is independent from one another, their order is inconsequential; thus we pool all novel samples into one support set during evaluation. Therefore, we focus on a single-step scenario with varying numbers of base and novel classes.

%

To evaluate performance, we use the following metrics:
\setlist{nolistsep}
\begin{itemize}[noitemsep,topsep=0pt]
    \item \textbf{Base Class Recognition (BCR):} Accuracy on $\mathcal{D}^{(b)}_{test}$
    \begin{equation}
        \mathrm{BCR} := \mathrm{ACC}(f(\cdot); \mathcal{D}^{(b)}_{test});
        \label{eq:bcr}
    \end{equation}
    \item \textbf{Novel Class Recognition (NCR):} Accuracy on $\mathcal{D}^{(n)}_{query}$
    \begin{equation}
        \mathrm{NCR} := \mathrm{ACC}(f(\cdot); \mathcal{D}^{(n)}_{query});
        \label{eq:ncr}
    \end{equation}
    \item \textbf{Average Class Recognition (AVG):} 
    Average accuracy defined as $\mathrm{AVG} := (\mathrm{BCR}+\mathrm{NCR})/2$;
\end{itemize}
where $\mathrm{ACC}(f(\cdot); \mathcal{D})$ is the accuracy of model $f(\cdot)$ on data $\mathcal{D}$. A common strategy in FSCIL is the use of a set of (\textit{empirical}) prototypes, $\mathcal{C} = \{ \mathbf{c}_{k}\}_{k \in \mathcal{K}}$ to serve as class representatives in the feature space and are defined as the centroid of the class-wise features: 
\begin{equation}
    \mathbf{c}_k := \frac{1}{|F_{k}|} \sum_{\mathbf{x}\in F_{k}}f(\mathbf{x}), \quad  \forall k \in \mathcal{K},
    \label{eq:proto}
\end{equation}
where $|\cdot|$ denotes the cardinality operator. 
For base classes $k_b\in\mathcal{K}_b$, $F_{k_b}$ consists of training samples for class $k_b$ from $X^{(b)}_{train}$, hence empirical prototypes are computed using a large training set and are good approximations of the \textit{true} prototypes $\boldsymbol{\mu}_{k_b}$, i.e., $\mathbf{c}_{k_b} \approx \boldsymbol{\mu}_{k_b}$. 
For novel classes in the 1SCIL setup $k_n\in\mathcal{K}_n$, instead, $F_{k_n}$ consists of the unique support sample for class $k_n$ from $X^{(n)}_{support}$, hence, empirical prototypes are computed using the single provided sample. 


For the experiments, we employ an episodic training framework \cite{li2019episodic}. Specifically, we sample a subset of classes $\mathcal{K}_b' \sim \mathcal{K}_b$, with each class $k'\in\mathcal{K}_b'$ contributing a support set $\mathcal{S}_{k'} \sim \mathcal{D}_{train}^{(b)}$ and a disjoint query set $\mathcal{Q}_{k'} \sim \mathcal{D}_{train}^{(b)}$. 
Empirical class prototypes $\mathbf{c}_{k'}$ are calculated as the mean embeddings of samples in $\mathcal{S}_{k'}$. 


\begin{figure*}[t]
    \centering
    \includegraphics[width=0.9\linewidth]{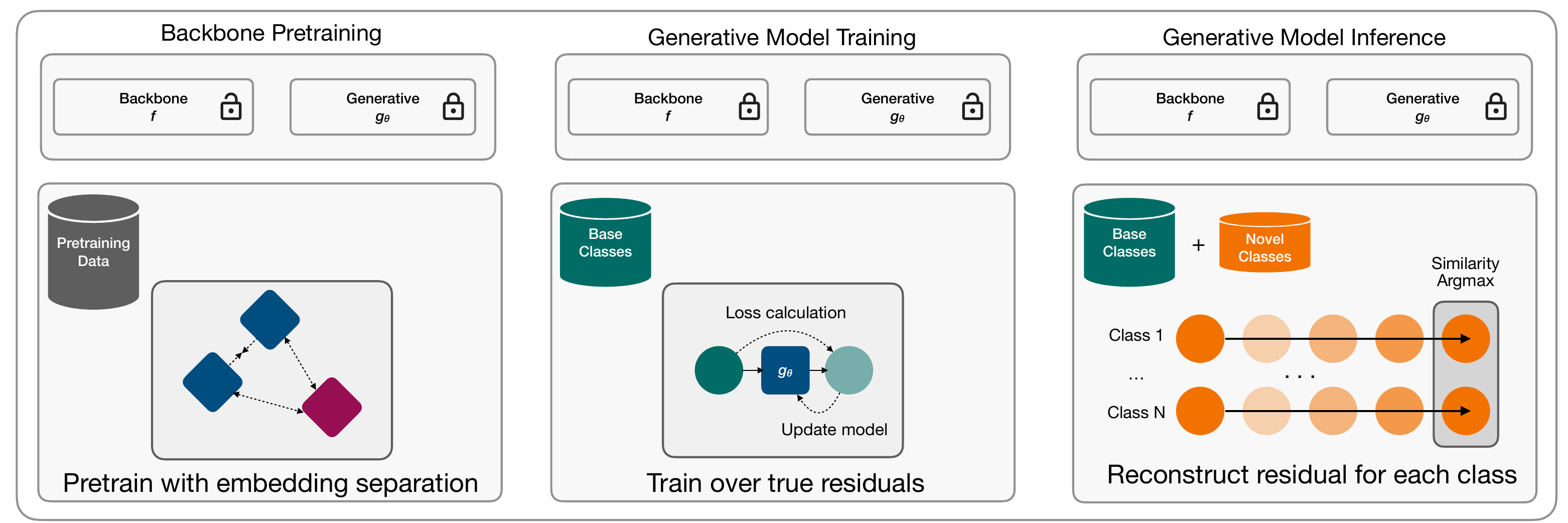}
\caption{{Overview of our \ours~method.} We propose to train a generative model (such as a VAE or a diffusion model) on the residual embedding distribution of base classes and use it as a prior to predict novel class distributions.}
\label{fig:method}
\end{figure*}

\section{Method}
\label{sec:method}

\subsection{Our Method}
\label{sec:method:ours}
\keypoint{Hypothesis}
This work is predicated on the notion that existing methods rely on weak priors for novel classes, thereby necessitating access to substantial novel class data for effective classification.
We hypothesize that novel class embeddings, under a well-generalized backbone, may exhibit structurally similar distributions to base class embeddings. 
Specifically, we posit that embedding distributions of base and novel classes may differ primarily in their means, while their intra-class structures remain consistent. We visualize this in Fig.~\ref{fig:setup} and we empirically explore it in Sec.~\ref{sec:results}. 
As discussed in Sec.~\ref{sec:intro}, this hypothesis arises from prior work that observes related classes have structural similarities in 
the embedding space. 

Formally, given a generic input $\mathbf{x}_k$ of class $k$, we assume 
\begin{equation}
%
%
    p \left( f ( \mathbf{x}_k ) | k=k_n \right) \approx p \left( f ( \mathbf{x}_k) | k=k_b \right) - \boldsymbol{\mu}_{k_b} + \boldsymbol{\mu}_{k_n},
\end{equation}
where the left-hand side represents the embedding distribution spanned by $f(\mathbf{x}_k)$ conditioned on the true class of $\mathbf{x}_k$ being $k_{n}$, and $\approx$ 
represents convergence in distribution.
In other words, changing the true class conditioning from $k_n$ to $k_b$ implies that the embedding distributions are related by a translation $\boldsymbol{\mu}_{k_n} - \boldsymbol{\mu}_{k_b}$.
Especially interesting for FSCIL is the case where $k_n\in\mathcal{K}_n$ and $k_b\in\mathcal{K}_b$.



\keypoint{Modeling Class Distributions}
Under the above mentioned hypothesis, we derive that class embedding distributions share a common structure, which we capture via a shared distribution $p(\mathbf{v})$ by:
\begin{equation}
    p \left( f(\mathbf{x}_k) \right) = \boldsymbol{\mu}_{k} + p(\mathbf{v}).
\end{equation}

Then, we construct class residual distributions by subtracting the respective prototype from class embedding distributions.
%
Provided that ${\mathbf{c}_{k} \approx \boldsymbol{\mu}_{k}}, \forall k \in \mathcal{K}$, we derive that all class residual distributions are approximated by $p(\mathbf{v})$:
\begin{equation}
\label{eq:vk}
    p(\mathbf{v}_{k}) = p \left( f(\mathbf{x}_k) \right) - \mathbf{c}_{k} = (\boldsymbol{\mu}_{k} + p(\mathbf{v})) - \mathbf{c}_{k} \approx p(\mathbf{v}).
\end{equation}

Hence, our objective is to construct a generative model $g_{\theta}(\cdot)$ to learn the shared distribution of the residuals, $p(\mathbf{v})$, from the base classes and use it for modeling the residual distribution of novel classes.


\textit{Non-Conditioned Case:} Eq.~\eqref{eq:vk} implies that distributions of residual embeddings are nearly independent of the class, i.e., ${p(\mathbf{v}_{k_1}) \approx p(\mathbf{v}_{k_2}) \approx p(\mathbf{v}), \forall k_1 \in \mathcal{K},k_2 \in \mathcal{K}}$. Hence, a generative model capturing the shared residual distribution $p(\mathbf{v})$ can also generalize to novel classes, and we can identify the class whose residual maximizes similarity with the learned distribution. 
In this case, the residual is the only input to the generative model: $g_\theta(\mathbf{v}_k)$. We refer the reader to the Appendix 
for results on this case.

%

\textit{Conditioned Case:}  In the 1SCIL setting, however, novel class prototypes are a weak approximation of the true prototypes. To cope with this, we empirically found that providing the generative model with the candidate class prototype $\mathbf{c}_{k}$ as context alongside the residual $\mathbf{v}_{k}$ allowed the model to improve reconstruction. In this case we write $g_\theta(\mathbf{v}_k, \mathbf{c}_k)$.


In our work, we implement $g_\theta(\cdot)$ as a VAE or a diffusion model, making no assumptions about the specific form of the embedding distributions. 
We assume some similarity across classes but do not require distributions to be Gaussian or unimodal. 
This 
allows our method to handle the diverse and intricate structures found in the embedding space, offering a significant advantage over existing methods.

\subsection{Training and Inference Overview}

\keypoint{Base Training of the Backbone $f(\cdot)$}
The generative model described above is agnostic to the choice of base training pipeline for the backbone model, provided it ensures good inter-class separation and similar structural properties across class embeddings.




\begin{algorithm}[t]
\caption{Base Training of the Generative Model}\label{alg:training}
\begin{algorithmic}
\REQUIRE base dataset $\mathcal{D}_{train}^{(b)}$, $N$ episodes, $\mathcal{K}_{b}$ classes

\FOR{$N$ episodes}
\STATE $\mathcal{K}^{'}_{b} \sim \mathcal{K}_{b}$  \COMMENT{sample classes for episode}
\STATE $\mathcal{Q}_{k'}, \mathcal{S}_{k'} \sim \mathcal{D}_{train}^{(b)}, \forall k'\in\mathcal{K}^{'}_{b}$ \COMMENT{query and support set} 
\STATE $\mathbf{c}_{k} \gets \frac{1}{|\mathcal{S}_{k} |}\sum_{\mathbf{x} \in \mathcal{S}_{k}} f(\mathbf{x}) \; \forall k \in \mathcal{K}$ \COMMENT{class prototypes}
\FOR{ \textbf{each} ($\mathbf{x}_i, k')$ in $\mathcal{Q}_{k'}, \ \forall k'\in\mathcal{K}_b$ }

\STATE $\mathbf{v}_{k'} \gets f(\mathbf{x}_i) - \mathbf{c}_{k'}$ \COMMENT{subtract prototypes}
\STATE $\hat{\mathbf{v}}_{k'} \gets g_{\theta}(\mathbf{v}_{k'}, \mathbf{c}_{k'})$ \COMMENT{get generative model output}
\STATE Compute loss $\mathcal{L}$ (Eqs.~\eqref{eq:my_vae_loss}-\eqref{eq:my_diffusion_loss}) and backpropagate
\ENDFOR
\ENDFOR
\end{algorithmic}
\end{algorithm}

\keypoint{Base Training of the Generative Model}
For the 1SCIL setup, we aim to leverage the generative model to learn the shared residual embedding distribution conditioned on true class prototypes  (Algorithm~\ref{alg:training}).
For each sample in the episode ${(\mathbf{x}_{i}, k') \in \mathcal{Q}_{k'}}$, the generative model $g_{\theta}(\cdot)$ takes as input the residual vector $\mathbf{v}_{i} = \mathbf{x}_{i} - \mathbf{c}_{k'}$, and the current support-set prototype $\mathbf{c}_{k'}$.
At the end of training, class mean vectors are calculated over the entirety of $\mathcal{D}_{train}^{(b)}$ and stored as class prototypes (left side of Fig.~\ref{fig:method}).


\keypoint{Few-Shot Sessions} The model is frozen during few-shot sessions. Instead, prototypes for novel classes are computed and stored, as defined in Eq.~\eqref{eq:proto}. The generative model $g_{\theta}(\cdot)$ is not used at this stage.

\keypoint{Inference Stage}
During inference, standard FSCIL methods assign a query sample $\mathbf{x}_{\bar{k}} \in (\mathcal{D}^{(n)}_{query} \cup \mathcal{D}_{test}^{(b)})$ with true class $\bar{k}\in\mathcal{K}$ to the class $\hat{k}$ whose prototype is closest in feature space, based on a chosen metric (e.g., cosine or Euclidean). 
Our approach follows a similar rationale but utilizes residuals and involves three steps (see right side of Fig.~\ref{fig:method} and  Algorithm~\ref{alg:inference}):
\begin{itemize}
    \item \textit{Residual Calculation.} 
    For each candidate class $k\in\mathcal{K}$, we compute the residual embedding $\mathbf{v}_{k} = f(\mathbf{x}_{\bar{k}}) - \mathbf{c}_{k}$.
    \item \textit{Similarity Computation.} Then, we get the predicted residual via the generative model $\hat{\mathbf{v}}_{k}=g_{\theta}(\mathbf{v}_{k}, \mathbf{c}_{k}), \forall k\in\mathcal{K}$ and compute the similarity between the true and predicted residual. The similarity metrics used here are given in Eqs.~\eqref{eq:vae_inf} for VAE and \eqref{eq:diff_inf} for diffusion model.
    \item \textit{Classification Criteria.} The generator is trained only on correct-class residuals, thus it learns to reconstruct examples from the true residual distribution. 
    At inference, the correct class residual yields high similarity (low reconstruction error), while 
    incorrect classes have low similarity. 
    Therefore, the class $\hat{k}$ maximizing the similarity between true and predicted residual is selected as output.
\end{itemize}



\subsection{VAE Implementation}

VAEs are generative models that maximize the likelihood of observed data $p(\mathbf{v})$. 
VAEs assume that observed data is generated by some latent variable $\mathbf{z}$, modeled through an encoder $q_\phi(\mathbf{z}|\mathbf{v})$.  A decoder then learns the distribution of the observed data over the latent variables $p_{\psi}(\mathbf{v}|\mathbf{z})$.

Since directly computing the likelihood $p(\mathbf{v})$ is intractable, VAEs maximize Evidence Lower Bound 
instead:
\begin{equation}
    \log\, p(\mathbf{v}) \! \geq \! \underbrace{\mathbb{E}_{q_{\phi} \!(\mathbf{z}|\mathbf{v})} \left[ \log\,p_{\psi}(\mathbf{v}| \mathbf{z}) \right]}_{\text{reconstruction loss}} 
    - \underbrace{\mathcal{D}_{KL}(q_{\phi}(\mathbf{z}|\mathbf{v})\: || \: p(\mathbf{z})).}_{\text{prior matching term}} \label{eq:elbo}
\end{equation}

VAEs can be extended to conditional settings, where auxiliary information, in our case a prototype $\mathbf{c}\in\mathcal{C}$, is used to model conditional distributions $q_{\phi}(\mathbf{z}|\mathbf{c}, \mathbf{v})$ and $p_{\psi}(\mathbf{v}|\mathbf{c}, \mathbf{z})$.

\keypoint{Base Training Session for VAE}
To train the VAE to reconstruct the residual for each sample in $(\mathbf{x}_i,k')\in\mathcal{Q}_{k'}$, we use the following objective:
    \begin{equation}
    \label{eq:my_vae_loss}
        \mathcal{L}_{\text{vae}} \!\! =\!\! \underbrace{-\mathbb{E}_{q_{\!\phi} \!(\mathbf{z}|\mathbf{v}_{\!k'}\!, \mathbf{c}_{\!k'}\!)} \! \! \left[ \log p_{\!\psi}\!(\!\mathbf{v}_{\!k'}\! | \mathbf{z} , \!\mathbf{c}_{\!k'}\!) \!\right]}_{\text{reconstruction term}} \! + \! \underbrace{\mathcal{D}_{K\!L}\!(\!q_{\phi} \!(\mathbf{z}|\mathbf{v}_{\!k'}, \! \mathbf{c}_{\!k'}\!) \! || \! \: p(\!\mathbf{z}\!)\!)}_{\text{prior matching term}}\!.
    \end{equation}
Simply, the model attempts to reconstruct the given residual, with additional context being provided in the form of the relevant class prototype (refer to \textit{Conditioned Case} in Sec.~\ref{sec:method:ours}).\\
\keypoint{VAE Similarity Computation}
During inference, we calculate the similarity between the true and the per-class predicted residual via the log-likelihood lower bound:
    \begin{multline}\label{eq:vae_inf}
        \text{score}({k})= \mathbb{E}_{q_{\phi}(\mathbf{z}|\mathbf{v}_{k}, \mathbf{c}_{k})} \left[ \log\,p_{\psi}(\mathbf{v}_{k}| \mathbf{z}, \mathbf{c}_{k}) \right]\\
        - \mathcal{D}_{KL}\left[ q_{\phi}(\mathbf{z}|\mathbf{v}_{k}, \mathbf{c}_{k})\: || \: p(\mathbf{z}) \right], \forall k\in \mathcal{K}.
    \end{multline}


\begin{table*}[!t]
\small
\centering
\setlength{\tabcolsep}{4pt}
\begin{tabular}{l ccc ccc ccc ccc}
\toprule
& \multicolumn{6}{c}{\textbf{DINOv2-s}} & \multicolumn{6}{c}{\textbf{ResNet18}} \\
\cmidrule(lr){2-7} \cmidrule(lr){8-13}
\textbf{Method} & \multicolumn{3}{c}{\textbf{CORe50}} & \multicolumn{3}{c}{\textbf{iCubWorld}} & \multicolumn{3}{c}{\textbf{CUB200}} & \multicolumn{3}{c}{\textbf{CIFAR100}} \\
\cmidrule(lr){2-4} \cmidrule(lr){5-7} \cmidrule(lr){8-10} \cmidrule(lr){11-13}
& \textbf{BCR} & \textbf{NCR} & \textbf{AVG} & \textbf{BCR} & \textbf{NCR} & \textbf{AVG} & \textbf{BCR} & \textbf{NCR} & \textbf{AVG} & \textbf{BCR} & \textbf{NCR} & \textbf{AVG} \\
\midrule
ProtoNet \cite{snell2017prototypical} & 85.4 & 60.7 & 73.1 & 97.9 & 68.3 & 83.1 & \textbf{74.1} & 26.2 & 50.2 & 76.1 & 7.3 & 41.7 \\
RelationNet \cite{sung2018learning} & 99.5 & 81.3 & 90.4 & \textbf{100} & 67.2 & 83.6 & 73.1 & 21.9 & 47.5 & 50.1 & 3.0 & 26.6 \\
FACT \cite{zhou2022forward} & 97.8 & 17.7 & 57.8 & 99.6 & 76.7 & 88.2 & 70.6 & 23.5 & 47.0 & \textbf{84.8} & 17.3 & 51.0 \\
LIMIT \cite{zhou2023few} & 72.8 & 18.9 & 45.8 & 97.3 & 52.1 & 74.7 & 58.8 & 14.6 & 36.7 & 76.2 & 19.4 & 47.8 \\
SAVC \cite{song2023learning} & 80.8 & 13.0 & 46.9 & 96.8 & 80.7 & 88.8 & 70.6 & 28.2 & 49.4 & 84.7 & 22.3 & \textbf{53.5} \\
OrCo \cite{ahmed2024orco} & \textbf{99.8} & 30.2 & 65.0 & \textbf{100} & 69.9 & 85.0 & 71.6 & 12.5 & 42.0 & 78.8 & 14.9 & 46.8 \\
SLDA-fixed-$\Sigma$ \cite{hayes2020lifelong} & 95.2 & 35.7 & 65.5 & \textbf{100} & 23.5 & 61.8 & 73.7 & 6.8 & 40.3 & 75.4 & 4.3 & 39.9 \\
SLDA \cite{hayes2020lifelong} & 95.4 & 36.1 & 65.8 & \textbf{100} & 23.7 & 61.2 & 73.9 & 0.3 & 37.2 & 75.9 & 1.3 & 38.6 \\
SimpleShot \cite{wang2019simpleshot} & 86.6 & 62.9 & 74.8 & 97.8 & 71.8 & 84.8 & 73.7 & 29.5 & 51.6 & 75.8 & 11.8 & 43.8 \\
\midrule
\textbf{\ours-VAE (ours)} & 87.1 & 86.7 & 86.9 & 97.9 & 83.6 & 90.8 & 73.7 & \textbf{46.6} & \textbf{60.2} & 74.7 & \textbf{23.3} & 49.0 \\
\textbf{\ours-Diffusion (ours)} & 88.2 & \textbf{93.0} & \textbf{90.6 }& 99.0 & \textbf{91.3} & \textbf{95.2} & 73.0 & 41.5 & 57.2 & 74.2 & 19.6 & 46.9 \\
\bottomrule
\end{tabular}%
\caption{{Single-class 1SCIL results.} Our approach consistently achieves the highest NCR.}
\label{tab:1sn1ncn}
\end{table*}

\begin{table*}[!t]
\small
\centering
\setlength{\tabcolsep}{4pt}
\begin{tabular}{l ccc ccc ccc ccc}
\toprule
& \multicolumn{6}{c}{\textbf{DINOv2-s}} & \multicolumn{6}{c}{\textbf{ResNet18}} \\
\cmidrule(lr){2-7} \cmidrule(lr){8-13}
\textbf{Method}& \multicolumn{3}{c}{\textbf{CORe50}} & \multicolumn{3}{c}{\textbf{iCubWorld}} & \multicolumn{3}{c}{\textbf{CUB200}} & \multicolumn{3}{c}{\textbf{CIFAR100}} \\
\cmidrule(lr){2-4} \cmidrule(lr){5-7} \cmidrule(lr){8-10} \cmidrule(lr){11-13}
& \textbf{BCR} & \textbf{NCR} & \textbf{AVG} & \textbf{BCR} & \textbf{NCR} & \textbf{AVG}& \textbf{BCR} & \textbf{NCR} & \textbf{AVG}& \textbf{BCR} & \textbf{NCR} & \textbf{AVG} \\ 
\midrule
ProtoNet \cite{snell2017prototypical} & 82.3 & 41.5 & 61.9 & 97.9 & 62.7 & 80.3 & \textbf{72.2} & 20.0 & 46.1 & 75.4 & 12.7 & 44.1\\ 

RelationNet \cite{sung2018learning} & 98.1 & 37.1 & 67.6 & 99.9 & 44.1 & 72.0 & 62.4 & 14.6 & 38.5 & 71.0 & 15.6 & 43.3 \\
FACT \cite{zhou2022forward} & 97.3 & 13.9 & 55.6 & 99.3 & 68.3 & 83.8 & 69.5 & 18.0 & 43.8 & \textbf{82.8} & 13.3 &\textbf{ 48.1}\\
LIMIT \cite{zhou2023few} & 69.7 & 11.1 & 40.4 & 96.1 & 37.8 & 67.0 & 58.3 & 10.2 & 34.3 & 75.2 & 15.1 & 45.2\\
SAVC \cite{song2023learning} & 76.6 & 18.2 & 47.4 & 95.6 & 78.0 & 86.8 & 69.4 & 18.4 & 43.9 & 82.7 & 12.8 & 47.8\\
OrCo \cite{ahmed2024orco} & \textbf{99.1} & 18.9 & 59.0 & \textbf{100} & 59.7 & 79.9 & 71.5 & 15.6 & 43.6 & 75.7 & 8.1 & 41.9\\
SLDA-fixed-$\Sigma$ \cite{hayes2020lifelong} & 95.2 & 27.1 & 61.2 & \textbf{100} & 32.0 & 66.0 & 73.0 & 7.5 & 59.9 & 75.5 & 5.1 & 61.4 \\
SLDA \cite{hayes2020lifelong} & 95.4 & 27.9 & 61.7 & \textbf{100} & 32.8 & 66.4 & 73.9 & 0.2 & 37.1 & 75.8 & 0.4 & 38.1 \\
SimpleShot \cite{wang2019simpleshot} & 83.4 & 42.7 & 63.1 & 95.0 & 63.6 & 79.3 & 70.7 & 21.1 & 45.9 & 72.9 & 13.1 & 43.0 \\
\midrule
\textbf{\ours-VAE (ours)} & 88.2 & 54.2 & 71.2 & 98.0 & 68.0 & 83.0 & 71.0 & 25.6 & 48.3 & 71.0 & 15.7 & 43.4\\ 
\textbf{\ours-Diffusion (ours)} & 87.1 & \textbf{58.6} & \textbf{72.9} & 95.1 & \textbf{80.1} & \textbf{87.6} & 71.9 & \textbf{28.7} & \textbf{50.3} & 71.4 & \textbf{16.0} & 43.7 \\
\bottomrule
\end{tabular}%
\caption{{Multi-class 1SCIL results.} 25 novel classes for DINOv2-s and 40 novel classes for ResNet18.}
\label{tab:1sn40ncn}
\end{table*}


\begin{algorithm}[t]
\caption{Inference with Generative Model}\label{alg:inference}
\begin{algorithmic}
\REQUIRE Class prototypes $\mathcal{C}$, query $\mathbf{x}_{\bar{k}} \in (\mathcal{D}^{(n)}_{query} \cup \mathcal{D}_{test}^{(b)})$

\FOR{$\mathbf{c}_{k}$ in $\mathcal{C}$}
\STATE $\mathbf{v}_{k} \gets f(\mathbf{x}_{\bar{k}}) - \mathbf{c}_{k}$ \COMMENT{subtract candidate prototype}
\STATE $\hat{\mathbf{v}}_{k} \gets g_{\theta}(\mathbf{v}_{k}, \mathbf{c}_{k})$ \COMMENT{get generative model output}
\STATE \textbf{store} score($\mathbf{v}_{k}$) \COMMENT{compute similarity, Eqs.~\eqref{eq:vae_inf}-\eqref{eq:diff_inf}}

\ENDFOR
\
\STATE \textbf{return} $\hat{k} = \argmax_{k\in\mathcal{K}} \text{sim}(\hat{\mathbf{v}}_k, \mathbf{v}_{k})$
\end{algorithmic}
\end{algorithm}

\subsection{Diffusion Implementation}

Diffusion models map a Gaussian noise distribution to a target distribution, and comprise a forward and reverse process.
The forward (encoder) process gradually corrupts data with Gaussian noise:
\begin{equation}
q(\mathbf{v}_{[1:T]}|\mathbf{v}_{[0]}) := \prod^{T}_{t=1} q(\mathbf{v}_{[t]} | \mathbf{v}_{[t-1]}),
\end{equation}
where $q(\mathbf{v}_{[t]} | \mathbf{v}_{[t-1]}) := \mathcal{N}(\mathbf{v}_{[t]}; \sqrt{1-\beta_{[t]}}\mathbf{v}_{[t-1]},\beta_{[t]} \mathbf{I})$, $\mathcal{N}$ is the normal distribution from which we draw $\mathbf{v}_{[t]}$ with a mean and covariance of $\sqrt{1-\beta_{[t]}}\mathbf{v}_{[t-1]}$ and $\beta_{[t]}\mathbf{I}$, respectively. $\beta_{[t]}$ is the variance of the noise at timestep $t$. The reverse (decoder) process denoises the corrupted sample by 
\begin{equation}
    p_{\theta}(\mathbf{v}_{[0:T]}) := p(\mathbf{v}_{[T]})\, \prod^{T}_{t=1} p_{\theta}(\mathbf{v}_{[t-1]}|\mathbf{v}_{[t]}).
\end{equation}

During training, input samples are corrupted with noise $\epsilon$ and the decoder is trained to predict such added noise by
\begin{equation}
    \mathcal{L}_{\text{ddpm}} = \mathbb{E}_{\Tilde{\mathbf{x}}, \tau} || \epsilon - \epsilon_{\theta}(\Tilde{\mathbf{x}}, \tau)||^{2}_{2}, 
    \label{eq:diff_loss}
\end{equation}
where $\Tilde{\mathbf{x}}$ is the noised sample and $\tau$ is the denoising timestep.

\keypoint{Base Training Session for Diffusion} To train the Diffusion model to reconstruct the residual for each sample in $(\mathbf{x}_i,k')\in\mathcal{Q}_{k'}$, we first perturb the residuals with Gaussian noise ($\tilde{\mathbf{v}_{k'}}$ is a noised version of $\mathbf{v}_{k'}$), and the model predicts the added noise at each timestep $\tau$:
    \begin{equation}
    \label{eq:my_diffusion_loss}
        \mathcal{L}_{\text{ddpm}} = \mathbb{E}_{\Tilde{\mathbf{v}}_{k'}, \tau} || \epsilon - \epsilon_{\theta}(\Tilde{\mathbf{v}}_{k'}, \mathbf{c}_{k'}, \tau)||^{2}_{2}. 
    \end{equation}
    The denoising operation enables the diffusion model to capture the shared residual distribution $p(\mathbf{v})$. To improve training, we also provide the relevant prototype as additional context to the diffusion model (\textit{Conditioned Case} in Sec.~\ref{sec:method:ours}).

\keypoint{Diffusion Similarity Computation}
To estimate similarity between the true and the per-class denoised (predicted) residual, we use the negative denoising loss:
\begin{equation}\label{eq:diff_inf}
    \text{score}({k}) = - \mathbb{E}_{\Tilde{\mathbf{v}}_{k}, \tau} || \epsilon - \epsilon_{\theta}(\Tilde{\mathbf{v}}_{k}, \mathbf{c}_{k}, \tau)||^{2}_{2}, \forall k\in\mathcal{K} .
    \end{equation}


\section{Results}
\label{sec:experiments}

\subsection{Experimental Setup}

\keypoint{Approaches}
We compare with approaches separating embeddings of base classes during pre-training (e.g., ProtoNet, FACT, LIMIT, SAVC, OrCO, and RelationNet) as well as approaches modeling novel class embeddings using base class statistics (e.g., SLDA-fixed-$\Sigma$, SLDA, and SimpleShot).
%
%
Notably, all prior works assume per-class embeddings follow a Gaussian distribution. 
We exclude methods that train the model during FSCIL, as it is incompatible with our on-device 1-shot setting.
We note that the VAE and diffusion models, as well as the MSE heads in RelationNet, each consist of $\sim 1\text{M}$ parameters.


\keypoint{Models}
We use two backbones: a pretrained DINOv2-s \cite{oquab2023dinov2} and a ResNet18 \cite{he2016deep}.
For Gen1S, SLDA and SimpleShot, we trained the ResNet18 with Prototypical loss over the base class data \cite{snell2017prototypical}. 
For DINOv2-s, we skip the training stage proposed in some prior work (ProtoNet, FACT, LIMIT, SAVC, OrCO) as it empirically led to lower results.

\keypoint{Datasets}
We use the CUB200 \cite{wah2011caltech} and CIFAR100 \cite{krizhevsky2009learning} datasets, using 160 and 60 base classes, respectively, and 40 novel classes each. Since  DINOv2-s was pretrained on both datasets, we evaluated it on the CORe50 \cite{lomonaco2017core50} and iCubWorld \cite{fanello2013icub} datasets, with 25 base classes and 25 novel classes. 
For all datasets, we also study the addition of a single novel class. 

\subsection{One-Shot Class Incremental Learning}
\label{sec:results}
Tables~\ref{tab:1sn40ncn} and~\ref{tab:1sn1ncn} report 1SCIL experiments, demonstrating that our method achieves state-of-the-art NCR across all datasets and models, regardless of whether the setting involves a single or multiple novel classes.
This highlights the efficacy of our approach in extracting valuable priors from base classes for downstream learning, making our method particularly advantageous for applications where novel classes are more important than base ones.
%
%
%
For three out of the four datasets, our method also achieves state-of-the-art average accuracy (AVG), striking a balance between maintaining BCR and improving NCR. 

Competing methods that focus on base embeddings separation at the base classes training stage can generally perform well on base classes but face challenges when requiring quick one-shot adaptation.
Other approaches that focus on improving few-shot sessions, like SLDA and SimpleShot, can achieve good NCR but are limited by the underlying assumption that novel class embeddings are Gaussian.

We observe that our method performs weaker when paired with ResNet18 compared to DINOv2-s. This suggests that DINOv2-s exhibits better generalization and greater similarity in class structures, making it more suited to our approach.

While our method performs best when paired with a diffusion model, the VAE-based version also consistently outperforms competing methods on NCR for all datasets except iCubWorld. This indicates that the method is robust and agnostic to the choice of the generative model.
As the number of novel classes increases, the diffusion model outperforms the VAE in both NCR and AVG. This is likely due to the diffusion model’s superior expressiveness and ability to learn complex, multi-modal distributions, which proves advantageous in scenarios with larger numbers of novel classes.

Last, 5-10 shots FSCIL results are shown in the Appendix, confirming that \ours~achieves the best NCR in most setups, although with reduced gains since competing methods can leverage the additional samples to improve accuracy.



\begin{figure}[t]
    \centerline{
        \includegraphics[width=1\linewidth]{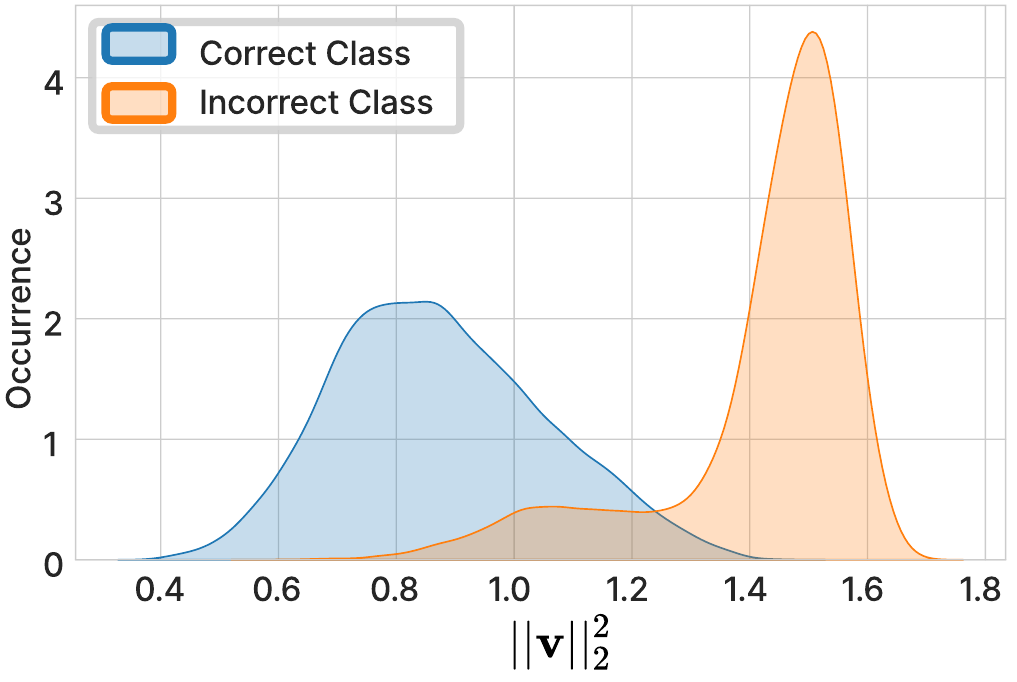}
    }
    \caption{{Distribution of L2 norm of residuals} for DINOv2-s on CORe50. While samples are typically closer to correct prototypes than to incorrect ones, the considerable overlap precludes the use of distance-based approaches.}
    \label{fig:density}
\end{figure}

\keypoint{Comparison to Distance-Based Methods}
Fig.~\ref{fig:density} shows the distribution of the L2 norm of the residuals from test samples to correct and incorrect class prototypes. 

We can draw two key observations.
First, the average distance in the embedding space of a sample to its positive class prototype is closer than its distance to any negative class prototype. This indicated that residuals can effectively distinguish between classes, as utilized in our method.
Second, despite the above, there is considerable overlap between the residual distributions of positive and negative classes. This complexity poses significant challenges for methods that rely on na\"ive distance metrics (e.g., Euclidean or cosine) to determine class membership (e.g., ProtoNet, SAVC, etc.).

\begin{table}[t]
\small
\centering
\setlength{\tabcolsep}{2.5pt}
\begin{tabular}{l l ccc}
\toprule
\textbf{Model} & \textbf{Dataset} & \textbf{Base ($\downarrow$)} & 
\textbf{Gauss ($\downarrow$)} 
& \textbf{Base vs.\ Gauss} \\
\midrule
DINOv2-s & CORe50    & 0.03 & 0.08 & 100\% \\
DINOv2-s & iCubWorld & 0.04 & 0.07 & 100\% \\
ResNet18 & CUB200    & 0.02 & 0.08 & 100\% \\
ResNet18 & CIFAR100  & 0.02 & 0.08 & 97.5\% \\
\bottomrule
\end{tabular}%
\caption{{Optimal transport of embedding residuals.} 
{Base:} Wasserstein distance between residuals of novel and base class embeddings.
{Gauss:} Wasserstein distance between residuals of novel class embeddings and a Gaussian fit on residuals of base class embeddings.
{Base vs.\ Gauss:} proportion of novel classes whose residuals  are closer to a base class than to the Gaussian fit on residuals of base classes.
A base class is almost always a better fit for each novel class than a Gaussian fit on base classes.}
\label{table:emb}
\end{table}

\begin{figure}[!t]
    \centerline{
        \includegraphics[ width=1\linewidth]{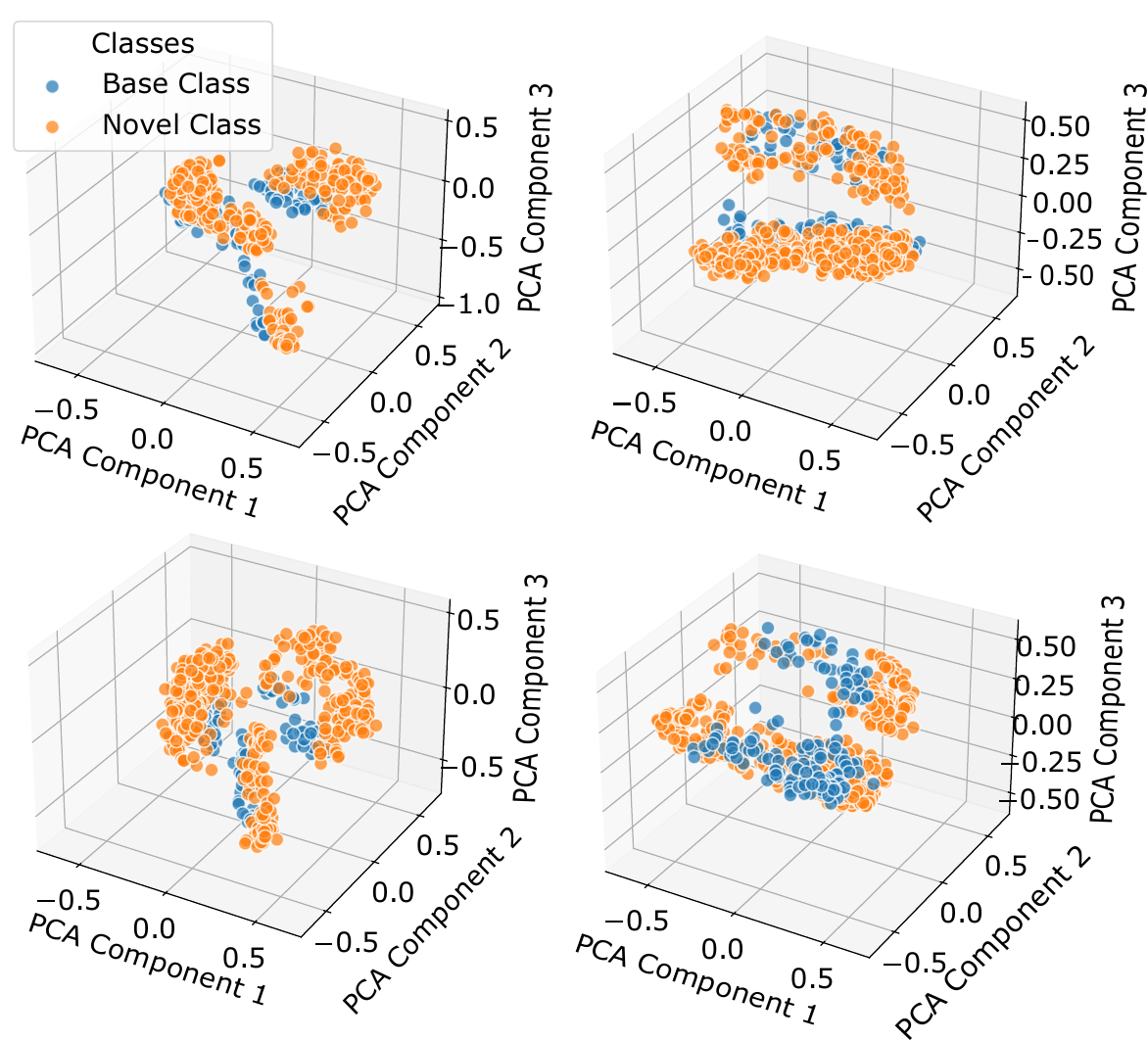}
    }
    \caption{{PCA of DINOv2-s residuals} of four CORe50 novel classes (orange) and corresponding base classes (blue) with similar structure. 
    Distributions are not Gaussian. The structure of base classes is often similar to that of novel classes, thus yielding an informative prior.}
    \label{fig:pca}
\end{figure}





\keypoint{Gaussian vs.\ Base Class Priors}  
Fig.~\ref{fig:pca} presents a PCA projection of four randomly picked novel class embeddings alongside a base class with similar structural properties. 
We appreciate qualitatively that base and novel classes exhibit similar embedding distributions. Such distributions are often multi-modal and rarely Gaussian, highlighting a key challenge for prototype-based FSCIL methods (e.g., ProtoNet and SLDA), which struggle to generalize in such scenarios.

To quantify this and further confirm our hypothesis that base classes provide informative priors for novel classes and that class embedding distributions are often not Gaussian, we compare the similarity of novel class residuals to that of the base classes, and to a Gaussian distribution that we fit over the base class residuals. We quantify the similarity via the formulation of an optimal transport problem. 
Specifically, for a set of base class embedding residuals $ A = \{\mathbf{a}_{1}, \mathbf{a}_{2}, \dots, \mathbf{a}_{M}\}$ and a set of novel class embedding residuals $ B = \{\mathbf{b}_{1}, \mathbf{b}_{2}, \dots, \mathbf{b}_{M}\}$, we measure the minimum work required to move $A$ onto $B$ using the Wasserstein distance: 
\begin{equation}
%
    W(A,B) = \inf_\pi \sqrt{\frac{1}{M} 
    \sum_{i=1}^{M} || \mathbf{a}_{i} - \mathbf{b}_{\pi(i)}||^{2}} ,
    \label{eq:wass}
\end{equation}
where the infimum is over all permutations $\pi$ of $M$ elements.

Table~\ref{table:emb} shows that $\geq97.5\%$ of times there exists a base class significantly more similar (in the embedding space) to a novel class than to the Gaussian model.
These findings confirm that base classes provide rich, informative priors beyond simple Gaussian assumptions, supporting our method.

\section{Conclusion}
\label{sec:conclusion}
In this work, we address the one-shot class-incremental learning (1SCIL) setup and introduce a novel approach that leverages a generative model capturing the residual distribution of class embeddings centered around their prototype. 
At inference time, the generative model is frozen and used as a prior for the structure of embeddings of novel classes.
%
Our method is agnostic to the choice of the generative model, although diffusion models proved to be the most effective.
%
%
While further advancements are needed to fully solve the challenges of 1SCIL, our method achieves state-of-the-art novel class recognition across all benchmarks as well as the highest average accuracy in three out of four benchmarks. 

\bibliography{main}


\newpage

\appendix

\renewcommand{\thefigure}{A\arabic{figure}}
\renewcommand{\theequation}{A\arabic{equation}}
\renewcommand{\thetable}{A\arabic{table}}

\setcounter{equation}{0}
\setcounter{figure}{0}
\setcounter{table}{0}

\section*{Appendix}

\noindent This appendix provides additional material to complement our main paper.
We start by presenting a visual intuition about residuals in Appendix~\ref{app:sec:residual}. 
We discuss the effect of conditioning generative models in Appendix~\ref{app:sec:conditioning}.
We present the influence of sampling on the diffusion model in Appendix~\ref{app:sec:sampling}.
Results with standard 5 or 10 shots are shown in Appendix~\ref{app:sec:FSCIL}.
Implementation details are shown in Appendix~\ref{app:sec:implementation}.
Finally, some limitations of our approach are discussed in Appendix~\ref{app:sec:limitations}.

\section{Intuition on Residuals}
\label{app:sec:residual}

Our Gen1S approach is based on the hypothesis that base and novel class embeddings have structural similarity that we capture via the residual space formed by subtracting class prototypes to embeddings of input samples.
A generative model is trained to learn the residual distribution, hence capturing the structure of base classes. 
At inference time, we select as output the class that minimizes the reconstruction error (i.e., maximizes similarity) between true and predicted residual.
We show an intuition in Figure~\ref{fig:setup}.

\section{Effect of Conditioning on Generative Models}
\label{app:sec:conditioning}

In the main paper, we discussed that we can employ generative models via prototype conditioning or without conditioning (see Section~4.1 of the main paper).

In the one-shot scenarios, the estimation error of the true prototype by means of the empirical prototype is notably high. 
To address this, we empirically found it beneficial to condition the diffusion model on the class prototype, enabling it to correct for this error effectively. 
As shown in Table~\ref{table:1shot_unc}, this conditioning step significantly improves performance compared to the unconditioned setup for both VAE and Diffusion model.

Interestingly, as shown in Appendix~\ref{app:sec:FSCIL}, the trend shifts as the number of shots increases. For higher shot numbers, the unconditioned model begins to outperform the conditioned model when using the DINOv2-s backbone. However, this is not the case for ResNet18, where the conditioned model continues to show better performance.

These findings suggest that conditioning helps to mitigate inter-class variations in embedding distributions, particularly when the backbone has limited generalization capabilities, as seen with ResNet18. Conversely, with better-generalized backbones like DINOv2-s, the unconditioned approach may become more effective as additional examples reduce the estimation error of the empirical prototype.

\begin{figure}[t]
    \centerline{
        \includegraphics[width=\linewidth]{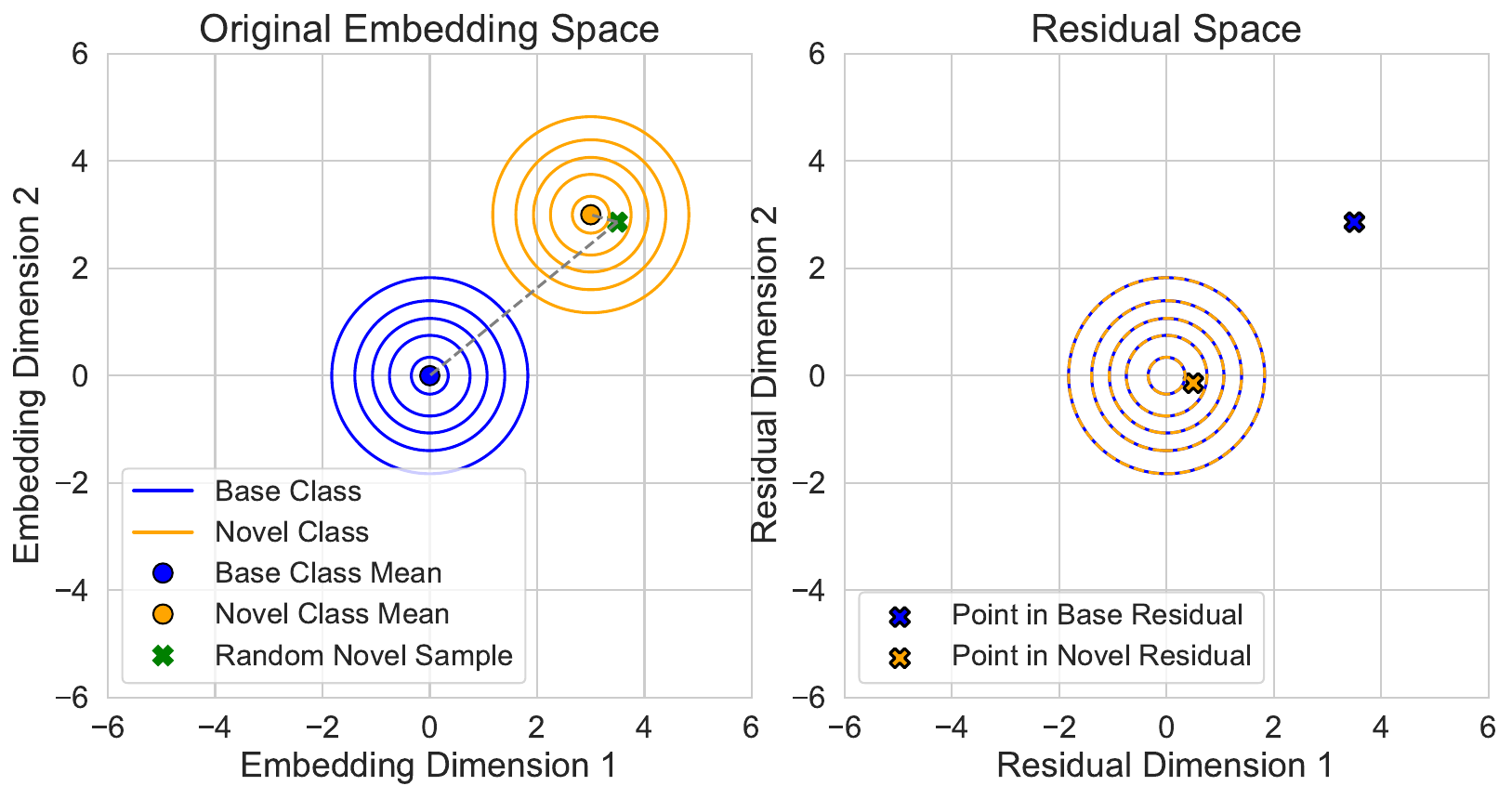}
    }

    \caption{Embeddings of different classes have similar distributions of residuals. 
    We map the original embeddings space (left plot) into a residual space by subtracting class prototypes (right plot). For visualization, we show the 2D space after PCA. We observe that base and novel class embeddings in FSCIL often exhibit similar intra-class structures. 
    We employ a generative model to learn the shared distribution of residuals across all base classes and predict it over the single support novel class sample.
    Classification on query novel class samples (green cross on the left plot) is performed by minimizing the reconstruction error between per-class predicted residual and true one. In the example shown, the yellow class would be selected as output (right plot).}
    \label{fig:setup}
\end{figure}

\begin{table*}[!ht]
\small
\centering
\setlength{\tabcolsep}{4pt}
\begin{tabular}{l ccc ccc ccc ccc}
\toprule
& \multicolumn{6}{c}{\textbf{DINOv2-s}} & \multicolumn{6}{c}{\textbf{ResNet18}} \\
\cmidrule(lr){2-7} \cmidrule(lr){8-13}
\textbf{Method}& \multicolumn{3}{c}{\textbf{CORe50}} & \multicolumn{3}{c}{\textbf{iCubWorld}} & \multicolumn{3}{c}{\textbf{CUB200}} & \multicolumn{3}{c}{\textbf{CIFAR100}} \\
\cmidrule(lr){2-4} \cmidrule(lr){5-7} \cmidrule(lr){8-10} \cmidrule(lr){11-13}
& \textbf{BCR} & \textbf{NCR} & \textbf{AVG} & \textbf{BCR} & \textbf{NCR} & \textbf{AVG}& \textbf{BCR} & \textbf{NCR} & \textbf{AVG}& \textbf{BCR} & \textbf{NCR} & \textbf{AVG} \\ 
\midrule
\textbf{\ours-VAE (ours, unconditioned)} & 80.0 & 41.5 & 60.8 & 96.9 & 60.9 & 78.9 & \textbf{74.0} & 18.9 & 46.5 & \textbf{75.7} & 12.7 & \textbf{44.2} \\
\textbf{\ours-VAE (ours, conditioned)} & \textbf{88.2} & \textbf{54.2} & \textbf{71.2} & \textbf{98.0} & \textbf{68.0} & \textbf{83.0} & 71.0 & \textbf{25.6} & \textbf{48.3} & 71.0 & \textbf{15.7} & 43.4\\
\midrule
\textbf{\ours-Diffusion (ours, unconditioned)} & 86.5 & 51.2 & 68.8 & \textbf{98.3} & 64.6 & 81.4 & 67.4 & 14.5 & 41.0 & \textbf{ 74.7} & 10.1 & 42.4\\ 
\textbf{\ours-Diffusion (ours, conditioned)} & \textbf{87.1} & \textbf{58.6} & \textbf{72.9} & 95.1 & \textbf{80.1} & \textbf{87.6} & \textbf{71.9} & \textbf{28.7} & \textbf{50.3} & 71.4 & \textbf{16.0} & \textbf{43.7} \\
\bottomrule
\end{tabular}%
\caption{{Conditioning effect on the generative model.} Multi-class one-shot FSCIL setup with 25 novel classes for DINOv2-s and 40 novel classes for ResNet18.}
\label{table:1shot_unc}
\end{table*}

\section{Diffusion Model: Influence of Sampling}
\label{app:sec:sampling}

\begin{figure}[!h]
    \centerline{
        \includegraphics[width=\linewidth]{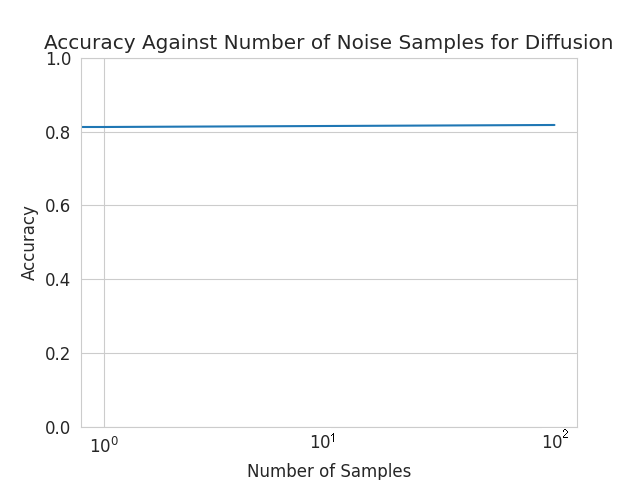}
    }
    \caption{{Influence of sampling on the diffusion model.} NCR accuracy of diffusion model for DINOv2-s iCubWorld at increasing number of noise samples during inference.}
    \label{fig:sampling}
\end{figure}

Figure~\ref{fig:sampling} illustrates how the performance of our diffusion-based generative model is influenced by the number of noise samples denoised during inference. Interestingly, increasing the number of noise samples has a minimal impact on the model's overall accuracy.

This observation highlights an important advantage of our method: even with sparse sampling, the diffusion model can maintain high classification accuracy. This property is particularly valuable in low-compute environments, as it allows for efficient inference without sacrificing performance.


\section{FSCIL Results}
\label{app:sec:FSCIL}

In this section, we analyze the performance of our approach in FSCIL scenarios with 5 or 10 support samples (\textit{shots}), considering both single-class (Tables~\ref{table:5shot_1c}, \ref{table:10shot_1c}) and multi-class (Tables~\ref{table:5shot_mc}, \ref{table:10shot_mc}) additions.

Our method consistently achieves competitive performance across all setups, surpassing state-of-the-art NCR scores in almost every cas, with only a few expections achieving marginally better results. 
This makes it particularly advantageous for applications where novel classes hold greater importance than base classes. 
Its effectiveness is most pronounced on DINOv2-s, benefiting from its rich and robust features.


\begin{table*}[!h]
\small
\centering
\setlength{\tabcolsep}{4pt}
\begin{tabular}{l ccc ccc ccc ccc}
\toprule
& \multicolumn{6}{c}{\textbf{DINOv2-s}} & \multicolumn{6}{c}{\textbf{ResNet18}} \\
\cmidrule(lr){2-7} \cmidrule(lr){8-13}
\textbf{Method} & \multicolumn{3}{c}{\textbf{CORe50}} & \multicolumn{3}{c}{\textbf{iCubWorld}} & \multicolumn{3}{c}{\textbf{CUB200}} & \multicolumn{3}{c}{\textbf{CIFAR100}} \\
\cmidrule(lr){2-4} \cmidrule(lr){5-7} \cmidrule(lr){8-10} \cmidrule(lr){11-13}
& \textbf{BCR} & \textbf{NCR} & \textbf{AVG} & \textbf{BCR} & \textbf{NCR} & \textbf{AVG} & \textbf{BCR} & \textbf{NCR} & \textbf{AVG} & \textbf{BCR} & \textbf{NCR} & \textbf{AVG} \\
\midrule
ProtoNet \cite{snell2017prototypical}  & 85.9 & 93.5 & 89.7 & 97.8 & 96.8 & 97.3 & {74.1} & 56.2 & 65.2 & 76.1 & 46.6 & 61.4 \\
RelationNet \cite{sung2018learning}  & 99.5 & 88.1 & \textbf{93.8} & \textbf{100} & 83.4 & 91.7 & 73.1 & 59.7 & 66.4 & 50.1 & 7.6 & 28.8 \\
FACT \cite{zhou2022forward}  & 97.8 & 34.0 & 65.9 & 99.6 & 91.0 & 95.3 & 70.6 & 57.1 & 63.8 & \textbf{84.7} & 44.8 & 64.8 \\
LIMIT \cite{zhou2023few}  & 72.9 & 12.6 & 42.8 & 97.2 & 60.7 & 79.0 & 58.8 & 44.7 & 51.8 & 76.1 & 42.2 & 59.2 \\
SAVC \cite{song2023learning}  & 80.9 & 46.8 & 63.8 & 96.8 & 94.5 & 95.6 & 70.6 & 52.3 & 61.4 & 84.6 & 52.8 & \textbf{68.7} \\
OrCo \cite{ahmed2024orco} & \textbf{99.8} & 42.4 & 71.1 & \textbf{100} & 91.8 & 95.9 & 71.6 & 46.4 & 59.0 & 78.8 & 28.1 & 53.4 \\
SLDA-fixed-$\Sigma$ \cite{hayes2020lifelong} & 95.3 & 80.0 & 87.7 & \textbf{100} & 93.5 & 96.8 & 73.7 & 30.6 & 52.2 & 75.5 & 18.9 & 47.2 \\
SLDA \cite{hayes2020lifelong} & 95.4 & 80.5 & 88.0 & \textbf{100} & 95.9 & 98.0 & 73.9 & 31.9 & 52.9 & 75.8 & 17.2 & 46.5 \\
SimpleShot \cite{wang2019simpleshot} & 86.3 & 94.6 & 90.5 & 98.4 & 97.1 & 97.8 & 74.0 & 60.7 & 67.4 & 76.4 & 45.3 & 60.9 \\
\midrule
\textbf{\ours-VAE (ours, unconditioned)} & 86.1 & 96.1 & 91.1 & 97.2 & 93.0 & 95.1 & \textbf{74.2} & 54.3 & 64.3 & 76.4 & 39.7 & 58.1 \\
\textbf{\ours-VAE (ours, conditioned)} & 87.1 & 93.2 & 90.2 & 98.0 & 93.7 & 95.8 & 73.7 & \textbf{73.5} & \textbf{73.6} & 74.7 & \textbf{49.1} & 61.9 \\
\textbf{\ours-Diffusion (ours, unconditioned)} & 87.8 & 85.5 & 86.6 & 98.7 & 94.7 & 96.7 & 67.6 & 41.2 & 54.4 & 75.1 & 42.7 & 58.9 \\
\textbf{\ours-Diffusion (ours, conditioned)} & 88.5 & \textbf{98.6} & 93.6 & 98.9 & \textbf{99.5} & \textbf{99.2} & 73.0 & 65.4 & 69.2 & 74.1 & 43.2 & 58.6 \\
\bottomrule
\end{tabular}%
\caption{{Single-class five-shot FSCIL results} with 1 novel class.}
\label{table:5shot_1c}
\end{table*}

\begin{table*}[!h]
\small
\centering
\setlength{\tabcolsep}{4pt}
\begin{tabular}{l ccc ccc ccc ccc}
\toprule
& \multicolumn{6}{c}{\textbf{DINOv2-s}} & \multicolumn{6}{c}{\textbf{ResNet18}} \\
\cmidrule(lr){2-7} \cmidrule(lr){8-13}
\textbf{Method} & \multicolumn{3}{c}{\textbf{CORe50}} & \multicolumn{3}{c}{\textbf{iCubWorld}} & \multicolumn{3}{c}{\textbf{CUB200}} & \multicolumn{3}{c}{\textbf{CIFAR100}} \\
\cmidrule(lr){2-4} \cmidrule(lr){5-7} \cmidrule(lr){8-10} \cmidrule(lr){11-13}
& \textbf{BCR} & \textbf{NCR} & \textbf{AVG} & \textbf{BCR} & \textbf{NCR} & \textbf{AVG} & \textbf{BCR} & \textbf{NCR} & \textbf{AVG} & \textbf{BCR} & \textbf{NCR} & \textbf{AVG} \\
\midrule
ProtoNet \cite{snell2017prototypical} & 82.8 & 67.0 & 74.9 & 96.5 & 89.6 & 93.0 & 73.5 & 49.6 & 61.6 & 74.4 & 36.5 & 55.4 \\
RelationNet \cite{sung2018learning} & 97.5 & 51.4 & 74.4 & 99.7 & 68.6 & 84.2 & 72.0 & 41.0 & 56.5 & 46.1 & 7.4 & 26.8 \\
FACT \cite{zhou2022forward}  & 97.0 & 30.5 & 63.8 & 99.1 & 84.2 & 91.6 & 69.0 & 43.4 & 56.2 & \textbf{80.5} & 30.9 & 55.7 \\
LIMIT \cite{zhou2023few}  & 70.0 & 13.1 & 41.6 & 96.1 & 46.7 & 71.4 & 57.9 & 34.3 & 46.1 & 73.6 & \textbf{39.7} & \textbf{56.6} \\
SAVC \cite{song2023learning} & 76.0 & 36.1 & 56.0 & 95.2 & 91.3 & 93.2 & 69.0 & 43.7 & 56.4 & \textbf{80.5} & 31.1 & 55.8 \\
OrCo \cite{ahmed2024orco} & \textbf{99.3} & 28.3 & 63.8 & 99.9 & 87.3 & 93.6 & 71.1 & 45.9 & 58.5 & 74.7 & 17.5 & 46.1 \\
SLDA-fixed-$\Sigma$ \cite{hayes2020lifelong} & 95.0 & 72.4 & 83.7 & 99.8 & 86.2 & 93.0 & 73.2 & 31.3 & 52.3 & 72.8 & 19.4 & 46.1 \\
SLDA \cite{hayes2020lifelong} & 95.4 & 73.3 & \textbf{84.4} & \textbf{100} & 89.0 & \textbf{94.5} & \textbf{73.9} & 28.0 & 51.0 & 74.3 & 16.4 & 45.4 \\
SimpleShot \cite{wang2019simpleshot} & 83.2 & 69.2 & 76.2 & 96.4 & 89.8 & 93.1 & 72.1 & 50.4 & 61.3 & 75.1 & 36.6 & 55.9 \\
\midrule
\textbf{\ours-VAE (ours, unconditioned)} & 82.8 & 66.5 & 74.7 & 96.7 & 87.6 & 92.2 & 73.6 & 50.4 & 62.0 & 74.6 & 36.1 & 55.4 \\
\textbf{\ours-VAE (ours, conditioned)} & 85.5 & \textbf{77.2} & 81.4 & 95.6 & 90.2 & 92.9 & 71.7 & 52.3 & 62.0 & 69.0 & 32.9 & 51.0 \\
\textbf{\ours-Diffusion (ours, unconditioned)} & 87.4 & 74.2 & 80.8 & 98.1 & 90.8 & 94.4 & 66.9 & 39.4 & 53.2 & 73.7 & 32.9 & 53.3 \\
\textbf{\ours-Diffusion (ours, conditioned)} & 83.8 & 77.1 & 80.4 & 94.1 & \textbf{93.6} & 93.8 & 71.5 & \textbf{52.7} & \textbf{62.1} & 69.4 & 36.6 & 53.0 \\
\bottomrule
\end{tabular}%
\caption{{Multi-class five-shot FSCIL results} with 25 novel classes for DINOv2-s and 40 novel classes for ResNet18.}
\label{table:5shot_mc}
\end{table*}

\begin{table*}[!h]
\small
\centering
\setlength{\tabcolsep}{4pt}
\begin{tabular}{l ccc ccc ccc ccc}
\toprule
& \multicolumn{6}{c}{\textbf{DINOv2-s}} & \multicolumn{6}{c}{\textbf{ResNet18}} \\
\cmidrule(lr){2-7} \cmidrule(lr){8-13}
\textbf{Method} & \multicolumn{3}{c}{\textbf{CORe50}} & \multicolumn{3}{c}{\textbf{iCubWorld}} & \multicolumn{3}{c}{\textbf{CUB200}} & \multicolumn{3}{c}{\textbf{CIFAR100}} \\
\cmidrule(lr){2-4} \cmidrule(lr){5-7} \cmidrule(lr){8-10} \cmidrule(lr){11-13}
& \textbf{BCR} & \textbf{NCR} & \textbf{AVG} & \textbf{BCR} & \textbf{NCR} & \textbf{AVG} & \textbf{BCR} & \textbf{NCR} & \textbf{AVG} & \textbf{BCR} & \textbf{NCR} & \textbf{AVG} \\
\midrule
ProtoNet \cite{snell2017prototypical} & 85.8 & 94.6 & 90.2 & 97.9 & 96.1 & 97.0 & {74.1} & 74.7 & 74.4 & 76.0 & 47.3 & 61.6 \\
RelationNet \cite{sung2018learning}  & 99.5 & 91.4 & \textbf{95.4} & \textbf{100} & 88.0 & 94.0 & 73.1 & 58.2 & 65.6 & 50.1 & 15.6 & 32.8 \\
FACT \cite{zhou2022forward} & 97.8 & 40.5 & 69.2 & 99.6 & 93.7 & 96.6 & 70.6 & 63.1 & 66.8 & \textbf{84.6} & 52.4 & \textbf{68.5} \\
LIMIT \cite{zhou2023few}  & 72.9 & 19.6 & 46.2 & 97.2 & 44.5 & 70.8 & 58.8 & 61.5 & 60.2 & 76.1 & \textbf{60.5} & 68.3 \\
SAVC \cite{song2023learning} & 80.8 & 46.9 & 63.8 & 96.7 & 94.5 & 95.6 & 70.5 & 67.8 & 69.2 & 84.5 & 47.8 & 66.2 \\
OrCo \cite{ahmed2024orco} & \textbf{99.9} & 49.7 & 74.8 & \textbf{100} & 92.1 & 96.0 & 71.6 & 69.3 & 70.4 & 78.8 & 31.3 & 55.0 \\
SLDA-fixed-$\Sigma$ \cite{hayes2020lifelong} & 95.4 & 83.2 & 89.3 & 99.8 & 92.0 & 95.9 & 72.4 & 55.7 & 64.1 & 74.4 & 43.1 & 58.8 \\
SLDA \cite{hayes2020lifelong} & 95.4 & 86.0 & 90.7 & \textbf{100} & 94.6 & 97.3 & 73.9 & 60.3 & 67.1 & 75.8 & 43.6 & 59.7 \\
SimpleShot \cite{wang2019simpleshot} & 86.2 & 94.8 & 90.5 & 98.0 & 97.5 & 97.8 & 74.0 & 76.2 & 75.1 & 76.1 & 49.6 & 62.9 \\
\midrule
\textbf{\ours-VAE (ours, unconditioned)} & 86.1 & 97.5 & 91.8 &  97.2 & 97.2 & 97.2 & \textbf{74.2} & 63.4 & 68.8 & 76.4 & 56.7 & 66.6 \\
\textbf{\ours-VAE (ours, conditioned)} & 87.1 & 97.7 & 92.4 & 98.0 & 98.8 & 98.4 & 73.7 & \textbf{80.2} & \textbf{77.0} & 74.6 & 55.9 & 65.2 \\
\textbf{\ours-Diffusion (ours, unconditioned)} & 87.9 & 93.7 & 90.8 & 98.7 & 98.0 & 98.4 & 67.6 & 56.1 & 61.8 & 75.0 & 48.9 & 62.0 \\
\textbf{\ours-Diffusion (ours, conditioned)} & 88.7 & \textbf{98.7} & 93.7 & 99.0 & \textbf{99.7} & \textbf{99.4} & 73.0 & 71.3 & 72.2 & 74.1 & 51.6 & 62.8 \\
\bottomrule
\end{tabular}%
\caption{{Single-class ten-shot FSCIL results} with 1 novel class.}
\label{table:10shot_1c}
\end{table*}

\begin{table*}[!h]
\small
\centering
\setlength{\tabcolsep}{4pt}
\begin{tabular}{l ccc ccc ccc ccc}
\toprule
& \multicolumn{6}{c}{\textbf{DINOv2-s}} & \multicolumn{6}{c}{\textbf{ResNet18}} \\
\cmidrule(lr){2-7} \cmidrule(lr){8-13}
\textbf{Method} & \multicolumn{3}{c}{\textbf{CORe50}} & \multicolumn{3}{c}{\textbf{iCubWorld}} & \multicolumn{3}{c}{\textbf{CUB200}} & \multicolumn{3}{c}{\textbf{CIFAR100}} \\
\cmidrule(lr){2-4} \cmidrule(lr){5-7} \cmidrule(lr){8-10} \cmidrule(lr){11-13}
& \textbf{BCR} & \textbf{NCR} & \textbf{AVG} & \textbf{BCR} & \textbf{NCR} & \textbf{AVG} & \textbf{BCR} & \textbf{NCR} & \textbf{AVG} & \textbf{BCR} & \textbf{NCR} & \textbf{AVG} \\
\midrule
ProtoNet \cite{snell2017prototypical} & 83.5 & 76.1 & 79.8 & 96.3 & 94.0 & 95.2 & 73.4 & \textbf{59.8} & \textbf{66.6} & 73.4 & 47.5 & 60.4 \\
RelationNet \cite{sung2018learning} & 97.5 & 56.1 & 76.8 & 99.6 & 72.8 & 86.2 & 72.1 & 47.6 & 59.8 & 45.8 & 8.5 & 27.2 \\
FACT \cite{zhou2022forward} & 96.7 & 38.4 & 67.6 & 99.1 & 87.1 & 93.1 & 68.8 & 51.1 & 60.0 & \textbf{79.4} & 37.8 & 58.6 \\
LIMIT \cite{zhou2023few} & 70.3 & 13.9 & 42.1 & 96.1 & 48.0 & 72.0 & 57.7 & 44.9 & 51.3 & 72.5 & \textbf{49.9} & \textbf{61.2} \\
SAVC \cite{song2023learning} & 75.6 & 42.5 & 59.0 & 94.9 & 93.0 & 94.0 & 68.8 & 51.2 & 60.0 & \textbf{79.4} & 37.6 & 58.5 \\
OrCo \cite{ahmed2024orco} & \textbf{99.4} & 30.7 & 65.0 & 99.9 & 91.1 & 95.5 & 70.9 & 55.7 & 63.3 & 74.4 & 21.3 & 47.8 \\
SLDA-fixed-$\Sigma$ \cite{hayes2020lifelong} & 95.4 & 75.9 & 85.7 & 99.8 & 90.1 & 95.0 & 71.5 & 48.7 & 60.1 & 74.2 & 34.8 & 54.5 \\
SLDA \cite{hayes2020lifelong} & 95.4 & \textbf{82.3} & \textbf{88.9} & \textbf{100} & 94.7 & \textbf{97.4} & \textbf{73.6} & 51.1 & 62.4 & 75.8 & 35.9 & 53.3 \\
SimpleShot \cite{wang2019simpleshot} & 84.2 & 77.3 & 80.8 & 95.3 & 94.1 & 94.7 & 73.3 & 59.7 & 66.5 & 73.6 & 45.8 & 59.7 \\
\midrule
\textbf{\ours-VAE (ours, unconditioned)} & 83.5 & 74.5 & 79.0 & 96.8 & 93.0 & 94.9 & 73.4 & 59.7 & \textbf{66.6} & 73.7 & 47.4 & 60.6 \\
\textbf{\ours-VAE (ours, conditioned)} & 85.6 & 79.9 & 82.8 & 95.4 & 93.1 & 94.2 & 71.6 & 58.2 & 64.9 & 68.0 & 40.8 & 54.4 \\
\textbf{\ours-Diffusion (ours, unconditioned)} & 87.4 & 81.1 & 84.2 & 98.0 & \textbf{95.1} & 96.6 & 66.6 & 48.9 & 57.8 & 72.7 & 44.3 & 58.5 \\
\textbf{\ours-Diffusion (ours, conditioned)} & 83.7 & 81.5 & 82.6 & 94.2 & 94.8 & 94.5 & 71.6 & 59.6 & 65.6 & 68.2 & 45.2 & 56.7 \\
\bottomrule
\end{tabular}%
\caption{{Multi-class ten-shot FSCIL results} with 25 novel classes for DINOv2-s and 40 novel classes for ResNet18.}
\label{table:10shot_mc}
\end{table*}

\section{Implementation Details}
\label{app:sec:implementation}

For the episodic setup, we set the batch size to 256, split across 8 classes for CORe50 and iCubWorld, 64 classes for CUB200, and 24 classes for CIFAR100, and with 2 samples per support set. 
We train for $N=51,200$ episodes.

For Diffusion, we train the model with $10,000$ denoising timesteps and during inference draw noise from $\tau=5,000$ and $\tau=100$, taking the average error. For VAE we sample 3 times and take the average error.

\subsection{Hardware and Software Details}
All experiments were run on an Ubuntu Unix-based machine with kernel version 22.04.5, equipped with 8 NVIDIA 3090 RTX GPUs (24GB, CUDA version 12.2, driver version 535.230.02), an Intel(R) Xeon(R) Gold 5218R CPU, Python 3.10.12, and 256GB of RAM. Training took approximately 8-10 hours per experiment. For inference, a single GPU is sufficient.



\section{Limitations}
\label{app:sec:limitations}
To the best of our knowledge, this work is the first to explore generative models in the most challenging one-shot class-incremental learning setup (1SCIL). 
By leveraging a generative model, our approach surpasses state-of-the-art performance, particularly in novel class recognition, enabling more adaptive pretrained models. This research bridges the gap between generative AI and few-shot learning, opening new directions in few-shot continual adaptation.

Our method shows strong real-world potential but has limitations. Despite a lightweight generative model ($\approx$1M parameters), requiring a forward pass for every class per each query may hinder real-time execution for large class counts. 
Future optimizations like class-wise batching or efficient sampling could improve inference time.
Additionally, our method relies on the quality of the embedding space generated by the backbone, and we do not propose techniques to enhance this, as it remains an active research area beyond the scope of this work. 
Severe distribution shifts between base and novel classes may also degrade performance, suggesting the need for domain adaptation generative models.

\end{document}